\documentclass{article}

\PassOptionsToPackage{numbers, compress}{natbib}

 \usepackage[final]{neurips_data_2023}

\usepackage[utf8]{inputenc} 
\usepackage[T1]{fontenc}    
\usepackage[hidelinks]{hyperref}       
\usepackage{url}            
\usepackage{booktabs}       
\usepackage{amsfonts}       
\usepackage{nicefrac}       
\usepackage{microtype}      
\usepackage{xcolor}         
\usepackage[bottom]{footmisc}

\newcommand{\draftonly}[1]{#1}
\renewcommand{\draftonly}[1]{}

\usepackage{times}
\usepackage{latexsym}
\usepackage{multirow}
\usepackage{enumitem}

\usepackage[T1]{fontenc}

\usepackage[utf8]{inputenc}

\usepackage{microtype}

\usepackage{amsmath}
\usepackage{amssymb}
\usepackage{siunitx}
\usepackage{booktabs}
\usepackage{multirow}
\usepackage{graphicx}

\usepackage{cancel}

\usepackage{listings}

\lstdefinestyle{simple_lst_style}{
    columns=flexible,
    keywordstyle=\color{red},
    numberstyle=\color{gray},
    stringstyle=\color{green},
    basicstyle=\ttfamily\small,
    identifierstyle=\color{black},
    commentstyle=\color{blue},
    breakatwhitespace=false,
    breaklines=true,
    captionpos=b,
    keepspaces=false,
    numbersep=5pt,
    showspaces=false,
    showstringspaces=false,
    showtabs=false,
    tabsize=2,
    frame=single,
}

\lstdefinelanguage{Prompt}{
    morekeywords={Human, Computer},
    sensitive=true,
    morestring=[b]",
}

\usepackage{float}
\usepackage{csquotes}

\newcommand\blfootnote[1]{%
  \begingroup
  \renewcommand\thefootnote{}\footnote{#1}%
  \addtocounter{footnote}{-1}%
  \endgroup
}

\newcommand{\ensuretext}[1]{#1}

\newcommand{\marker}[2]{\ensuremath{^{\textsc{#1}}_{\textsc{#2}}}}
\newcommand{\authorcomment}[4]{\draftonly{\ensuretext {\textcolor{#3}{[\marker{#1}{#2} #4]}}}}

\newcommand{\richard}[1]{\authorcomment{R}{S}{red!80}{#1}}
\newcommand{\samt}[1]{\authorcomment{S}{T}{purple!80}{#1}}

\newcommand{\jason}[1]{\authorcomment{J}{E}{orange!80}{#1}}
\newcommand{\adpauls}[1]{\authorcomment{A}{P}{green!80}{#1}}

\title{BenchCLAMP: A Benchmark for Evaluating Language Models on Syntactic and Semantic Parsing}

\author{%
Subhro Roy$^{1}$ \quad Sam Thomson$^{1}$ \quad Tongfei Chen$^{1}$ \quad Richard Shin$^{1}$ \\
\textbf{Adam Pauls}$^{2*}$ \quad \textbf{Jason Eisner}$^{1}$ \quad \textbf{Benjamin Van Durme}$^{1}$\vspace{0.03in} \\
$^{1}$Microsoft Semantic Machines \quad $^{2}$Scaled Cognition\vspace{0.03in}\\
\tt\small $^{1}$\{subhrroy,sathomso,tongfeichen,eush,jeisner,bevandur\}@microsoft.com\\
\tt\small $^{2}$adpauls@scaledcognition.com
}

\begin{document}

\maketitle

\begin{abstract}
Recent work has shown that generation from a prompted or fine-tuned language
model can perform well at semantic parsing when the output is constrained to be a valid 
semantic representation. 
We introduce {\bf BenchCLAMP}, a {\bf Bench}mark 
to evaluate {\bf C}onstrained {\bf LA}nguage {\bf M}odel {\bf P}arsing, 
that includes context-free grammars for seven semantic parsing datasets and two syntactic 
parsing datasets with varied output 
representations, as well as a constrained decoding interface to generate only valid outputs covered 
by these grammars.
We provide low, medium, and high resource splits for each dataset, allowing
accurate comparison of various language models under different data regimes.
Our benchmark
supports evaluation of language models using prompt-based learning as well as fine-tuning. We benchmark eight language models, including two GPT-3 variants available only through an API. Our experiments show
that encoder-decoder pretrained language models can achieve similar performance or 
surpass  
state-of-the-art methods for syntactic and semantic parsing when the model output is 
constrained to be valid.
\blfootnote{$\textbf{*}$ Work done while at Microsoft Semantic Machines.}

\end{abstract}

\section{Introduction}
Large pretrained language models can achieve state-of-the-art 
performance on a host of NLP tasks when fine-tuned on target data \cite{roberta-liu-2019, t5-raffel-2020, zero-label-wang-2021, deberta-v3-he-2021}.
Models like GPT-3 \cite{Brown:2020:gpt-3}, Codex \cite{codex-chen-2021} and T0 \cite{sanh2021multitask} have also shown 
impressive zero- and few-shot performance when prompted only with task descriptions and examples. 
Research on large language models is 
typically validated by performance on downstream NLP tasks. Past work has evaluated new pretrained
language models on classification, extraction, and 
generation, among others \cite{roberta-liu-2019, deberta-v3-he-2021, helm}. However, 
parsing tasks are generally not considered a testbed for such evaluation. The outputs 
of parsing tasks are structured objects such as parse trees or code. State-of-the-art
systems thus involve task- or dataset-specific model architectures and
target
representation constraints. Evaluating language models on parsing tasks test capabilities not captured 
by commonly used evaluation tasks.

Recently, \citet{shin-etal-2021-constrained} and \citet{scholak-etal-2021-picard} have shown that
generation from a fine-tuned or few-shot prompted language model
can perform competitively in
semantic parsing tasks, when the output of the language model is constrained to produce valid
meaning representations. However, it is still challenging to set up constrained generation for a 
new dataset and language model due to the variation in
output formalisms
and model-specific
tokenization. In this paper, we introduce a new benchmark called BenchCLAMP (Benchmark for 
Constrained Language Model Parsing)
that covers nine parsing datasets with seven different
\samt{\emph{meaning representations} %
same as above\ldots syntax parses aren't meaning representations.}
output formalisms.
We release context-free grammars for each dataset and provide a toolkit
to perform efficient constrained decoding to generate valid output representations\samt{same as above}.
Our
benchmark reduces
the barrier for language model developers to evaluate on parsing. The benchmark is
available at {\small\url{https://github.com/microsoft/semantic_parsing_with_constrained_lm}}.

We benchmark eight pretrained language models using BenchCLAMP. We find that fine-tuning encoder-decoder
pretrained language models can come close to or surpass the performance of state-of-the-art 
methods on all parsing datasets. Constrained generation via a domain-specific grammar provides 
performance gains for most fine-tuned language models. The improvement is high in low-resource 
settings but the relative improvement reduces when more training data becomes 
available. 
We find constrained decoding to be essential for few-shot prompted models and for tasks with 
complex constraints like constituency and dependency parsing. In both these cases, we find 
language
models struggle to generate valid representations without constrained decoding. In addition, 
we ablate different ways to encode context in the input, prompt structure, and retrieval methods 
for few-shot prompted language models. We present a comprehensive study establishing concrete 
techniques to reliably use language models for syntactic and semantic parsing tasks.

\section{Related Work}
\noindent
\textbf{Language Models for Semantic Parsing:} Recent work has shown that one can generate an analysis of a natural language sentence, such as a semantic parse, by asking a large language model to continue a prompt that includes the sentence \cite{codex-chen-2021,mtop-li-2021,schucher-2021}.
We refer to this as ``language model parsing.''
To avoid ill-formed analyses, it is possible to \emph{constrain} the generation so that the generated output satisfies hard task-specific constraints.  
\citet{shin-etal-2021-constrained} showed that constrained generation
from few-shot prompted GPT-3 and fine-tuned BART models outperformed task-specific semantic parsing architectures in
low-resource settings.
\citet{scholak-etal-2021-picard} were able to achieve state-of-the-art performance in SQL prediction by fine-tuning a T5-3B model \cite{t5-raffel-2020} and using constrained decoding.
Recent work on AMR parsing has also shown positive results with sequence-to-sequence training
with pretrained language model parameters \cite{cai-lam-2020-amr,bai-etal-2022-graph}.
As the above works used different evaluation settings, it is hard to see which techniques work best under different data regimes.

\noindent
\textbf{Language Models for Syntactic Parsing:}
Syntactic parsing tasks like constituency and dependency parsing requires the outputs to be
well aligned with the input. All input tokens need to be covered by the output constituency
or dependency parse. As a result, most solutions for syntactic parsing has involved 
custom decoders or inference algorithms \cite{kitaev-klein-2018-constituency, yang-tu-2022-bottom,zhang-etal-2017-stack,liu-zhang-transition-2017}. However there has been some
work on generating linearized representations of the output \cite{koo2015grammar,wiseman-rush-2016-sequence,li-etal-2018-seq2seq,fernandez-gonzalez-gomez-rodriguez-2020-enriched}. 

\noindent
\textbf{NLP Benchmarks:} Multiple benchmarks have been introduced to track progress on specific NLP tasks and to encourage multi-task learning with diverse datasets. The GLUE 
\cite{wang-etal-2018-glue}, SuperGLUE \cite{wang-etal-2019-superglue}, BIG-bench \cite{bigbench} and HELM \cite{helm} are 
widely used by language model developers.
However, these benchmarks focus on classification, span extraction and generation, 
and do not include
structured prediction tasks like parsing.
More recently, the UnifiedSKG \cite{UnifiedSKG} benchmark has been introduced that 
converts a suite of tasks requiring structured knowledge into text-to-text format. In contrast
to their work, we cover a wide range of parsing tasks covering AMR, SMCalFlow, constituency and
dependency parsing. While they focus on unconstrained generation, we develop grammars and 
constrained decoding interface to support valid representation generation.

\section{Benchmark Details}

\begin{table}[!htbp]
\caption{List of datasets covered by BenchCLAMP, along with evaluation metric and an example representation. All representations are linearized into a sequence that can be produced by
a language model. We use task and dataset specific metrics to evaluate performance.} 
\label{tab:datasets_covered}
\small
\centering
\begin{tabular}{@{}p{2.7cm} l p{8cm}}
\toprule
\bf Dataset & \bf Metric & \bf Example Representation \\
\midrule
SMCalFlow \cite{andreas-etal-2020-task} & \multirow{2}{*}{Lispress Match} & \texttt{(Yield (Event.start (FindNumNextEvent} \\
TreeDST \cite{cheng-etal-2020-conversational} & & \texttt{ (Event.subject\_? (?\textasciitilde = ``meeting'')) 1L)))}\\
\addlinespace

MTOP \cite{mtop-li-2021} & Exact Match & \texttt{[IN:Get\_Message [SL:Type\_Content video] [SL:Sender Atlas]]}\\
\addlinespace

Overnight \cite{wang-etal-2015-building} & Denotation Match & \texttt{(call listValue (call getProperty en.block.block1 (string color)))}\\
\addlinespace

Spider \cite{yu-etal-2018-spider} & Test suite & \multirow{2}{8cm}{\texttt{SELECT born\_state FROM head GROUP BY born\_state HAVING count(*) >= 3}}\\
CoSQL \cite{yu-etal-2019-cosql} & Execution & \\
\addlinespace
AMR 2.0 \cite{banarescu-etal-2013-AMR} & Smatch & \texttt{(e/establish-01 :ARG1 (m/model :mod (i/innovate-01 :ARG1 (i2/industry))))}\\
\midrule
PTB 3 \cite{marcus-etal-1993-ptb} (Constituency parsing) & EVALB & \texttt{(S (NP (NNP Ms.) (NNP Haag)) (VP (VBZ plays) (NP (NNP Elianti))) (. .))}\\
\addlinespace
UD-EWT \cite{univ-deps-211} (Dependency parsing) & LAS & \texttt{(1-Read, root, 0) (2-some, obj, 1) (3-of, case, 6) (4-the, det, 6) (5-following, amod, 6) (6-links, nmod, 2)}\\

\bottomrule
\end{tabular}
\end{table}

\subsection{Data Setup}
BenchCLAMP includes nine popular parsing datasets with a varied set of meaning representation formalisms (details in \autoref{tab:datasets_covered}). We create 
several splits for each dataset. For datasets that do not release public test sets, like SMCalFlow, Spider, and CoSQL, we treat the development set as the test set, and 
sample $10\%$ of the training set and treat it as the development set for splits creation.
\begin{enumerate}[leftmargin=*]
    \item We create \textbf{three low-resource train splits} of \num{500} examples, each 
    uniformly sampled from the
    training portion of the dataset. We create a single low-resource development set of \num{50} examples sampled from the development portion of the dataset. We report
    mean of these splits.\jason{does this mean the mean over 3 systems trained on the 3 train splits, with hyperparameters of each system tuned on the shared dev set}
    \item We similarly create \textbf{a medium-resource train split} of \num{5000} examples paired with 
    a development set of \num{500} examples.
    \item We consider \textbf{a high-resource split} with the entire training set of the dataset, paired with the medium-resource development set.
\end{enumerate}

To make it feasible for researchers to evaluate large pretrained models on BenchCLAMP, 
we randomly sample a smaller test set for datasets with large released test sets.
Specifically, we sample \num{2000} examples from the test sets of 
SMCalFlow, TreeDST and MTOP datasets and evaluate 
test performance on this smaller set. We use the full test set for all other datasets.
We also release a smaller randomly-sampled \num{100}-example test set for each dataset
to evaluate models accessed through costly API calls like GPT-3 and Codex. Results on a \num{100}-example test set will have wide error bars and should be used with caution.\jason{this is the reason to do sig tests in the comparisons.  Given these error bars, quite possible that a lot of the reported comparison results don't stand up?}
See the Appendix for a discussion on result variance. 
We allow for the evaluation on full test sets of the datasets to compare with state of the art results. 

For datasets that include dialogue interactions, we ensure that all turns of a dialogue belong to the 
same split. The Overnight train set was already small (< \num{5}k examples), so we do not
have a separate medium split for it. 
We release data splits for all domains of Overnight and all MTOP
languages, but for brevity we benchmark on a single Overnight domain (blocks) and a single MTOP 
language (English) in this paper. 

\noindent
\textbf{Linearizing Representations:} We use the dataset representations
as is for the MTOP, PTB-3 and the SQL datasets. 
For AMR, we use the setup provided by \cite{van-noord-bos-2017-dealing,van2017neural} to linearize the representations into sequences for training, and to convert output model
predictions to AMRs for evaluation. For SMCalFlow and TreeDST, we use the Lispress format (LISP-like serialization for programs) of the data released by \cite{platanios-etal-2021-value}. We linearize dependency parses into a sequence of dependency triples. For the example
in Table \ref{tab:datasets_covered}, our representation has the following form: 
\begin{displayquote}
\small
(Read, root, root) (some, obj, Read) (of, case, links) (the, det, links) (following, amod, links) (links, nmod, some) 
\end{displayquote}
To convert such a representation back to a dependency parse, we find head
token indices for each token based on string match with the predicted head token. In case there are
multiple mentions of the head token, we select the one that is closest
to the token being considered. We can correctly roundtrip $95.7\%$ of parses in the test set using this approach. For completion, we also release a loss less linearization
of dependency parses which is similar to the one above but with the head token replaced 
by the corresponding token index in the input sequence.

\subsection{Grammars}
We release context-free grammars for all datasets to constrain generation
to valid meaning representations. The grammar creation process is specific to
each dataset.
\begin{enumerate}[leftmargin=*]
    \item For SMCalFlow and TreeDST, we use the Lispress-format datasets
released by \citet{platanios-etal-2021-value}.\adpauls{I wonder if you need to use TreeDST-lisp or something to make very clear that this not the same
as the original TreeDST data. If one of the authors sees Table 1 with Lispress and a cite to their original paper (which also has some different LISP syntax), I think they will be very confused. } We create a non-terminal corresponding to each type
present in the training data. For each \mbox{(sub-)}expression with type $t$, we add 
a production rule with the non-terminal for $t$ generating the non-terminals of its component 
(sub-)expression types, or component terminal plan fragments. 

\item For MTOP, 
we add a non-terminal corresponding to each intent and slot. Each intent non-terminal generates an expression comprising slot non-terminals. Similarly each slot non-terminal can generate an
expression with nested intent non-terminals. Slot non-terminals can also generate terminal strings copied from the input utterance.

\item We use a publicly available SQL grammar \cite{antlr-2022} for Spider and CoSQL.
For each example, we  add schema-specific constraints to the grammar to generate
consistent table and column names. This is similar to ``parsing without guards'' in \citet{scholak-etal-2021-picard}.
\richard{You can say this is similar to ``parsing without guards'' in PICARD}

\item For constituency parsing, we define a non-terminal for an expression and a constituent 
label. We add production rules where the label non-terminal can produce any of the constituent 
labels seen in the training data. The expression non-terminal can either produce terminal tokens
or generate a constituent label coupled with a expression non-terminal. This context free grammar covers constituent parse tree representation shown in Table \ref{tab:datasets_covered}. 
To additionally ensure that all tokens in the input utterance are covered by the 
generated parse tree, we additionally maintain a state in our parsing algorithm during 
decoding, allowing
tokens to appear in the order seen in the utterance, and allowing the generation to end 
only when all input tokens have been generated.

\item For dependency parsing, we extract the set of dependency relations from the 
training set
and define a non-terminal that can produce any of them. We then define a non-terminal that
can generate a sequence of triples each comprising two tokens from the utterance and a dependency relation. Similar to our approach with constituency parsing, we maintain a parse state during decoding to ensure all tokens
from the utterance are covered in order in the generated output. 
\end{enumerate}

For all data splits, we use the full training data to derive the grammar\adpauls{I wonder if you should say "derive" instead of "induce", since you don't need any data to produce these grammars. The syntax of the language and type declarations are sufficient. Maybe a footnote would help?}. We envision that
in realistic scenarios, the grammar will be provided by a domain developer, and hence will have
complete coverage of the domain (even when some plan fragments might not have appeared in the
low-resource dataset). We also add results with grammars induced from low-resource 
splits in section \ref{subsec:low-resource-grammar}.

\section{Experimental Setup}
\subsection{Language Models Evaluated}
\label{subsec:lm-evaluated}
We use BenchCLAMP to fine-tune and evaluate five language models with varying number of parameters: T5-base (\num{220}M), T5-large (\num{770}M), T5-3B (\num{3}B) \cite{t5-raffel-2020},
BART-large (\num{406}M) \cite{lewis-etal-2020-bart} and CodeT5-base (\num{220}M) \cite{wang-etal-2021-codet5}.
The input to our model is
the utterance concatenated with the string representation of the context (conversation context, database schema, etc.), and the output is the target parse. 
We evaluate three large language models: GPT-3 \cite{Brown:2020:gpt-3}, Codex \cite{codex-chen-2021} and Llama-2 \cite{llama2}\footnote{We use \texttt{llama-2-7b} in our evaluation. Llama-2 is served using vLLM \cite{vllm} to create the same OpenAI API as GPT-based models.}, using few-shot prompting on the
$100$ example test sets. 
Unless otherwise state, we use the OpenAI API \texttt{text-davinci-001} for GPT-3
and \texttt{code-davinci-001} for Codex.
For each input (utterance concatenated with context) we select a set of $20$ relevant
examples from the training set using BM25 \cite{rubin-2021-in-context-learning} or
a sentence transformer \cite{reimers-2019-sentence-bert} based similarity model. We create a prompt using these
examples, following the template in \citet{shin-etal-2021-constrained} and limiting the total length of the prompt to be \num{1500} tokens. 
This leaves room in GPT-3's buffer to generate an output of up to \num{548} tokens. 
\samt{100 is \emph{really} small. Do we talk somewhere about how much we can trust these results? Like what are the error bars on accuracies on these test sets?}

We release data splits for all domains of Overnight and all languages in MTOP. But for brevity,
we benchmark on a single domain of Overnight (blocks) and a single language from MTOP (English).
All other datasets used in the benchmark are in English.
We evaluate few-shot prompted GPT-3, Codex, and Llama-2 on three BenchCLAMP datasets. All other models are evaluated on the complete evaluation suite.


We use the code released by \citet{shin-etal-2021-constrained} to support incremental constrained generation of 
semantic representations. This code maintains a chart according to Earley's algorithm
\cite{earley70} that can be used to determine the set of legal next tokens and can also be efficiently updated after a particular token is selected. 
We extend their method to support all autoregressive language models and sequence-to-sequence models.
Unless otherwise mentioned, we always use constrained decoding.

\begin{table}[!htbp]
\caption{Performance of language models on SMCalFlow, TreeDST and MTOP. $^\dagger$ indicates few-shot prompted and evaluated on the 100 example test set; numbers above the horizontal bar are thus not comparable to those below. $^1$ Prompt exemplars retrieved by BM25;  $^2$ by dense vector retrieval using SentenceT5-xxl.
Remaining LMs are finetuned and evaluated on BenchCLAMP test sets.
We report results with both constrained and unconstrained decoding to illustrate the contribution of constraints. Metrics are dataset-specific (see Table \ref{tab:datasets_covered}). The best score in each column is boldfaced.
\samt{not necessarily stat. significant, right?}
\jason{Can you say what the output formats are if they're not SQL?  And what numbers are being reported here?  What does boldface mean---are you boldfacing the best result in each column together with all results that are not significantly worse (I'd use a paired permutation test, $p < 0.05$, and am happy to give you the code snippet)?  Do Low, Med, and All refer to the amount of training and dev data used for the fine-tuning?  Should All be renamed to High for consistency with the text?}\jason{You should probably call out the difference between T5-large and T5-large unconstrained.  Is that the \emph{only} case where you measure the effect of the constraints?  It looks like the constraints sometimes help and sometimes hurt slightly.  I don't fully understand why---could you perhaps report the rate at which the 1-best answer of the unconstrained model fails the grammar?  We were talking on a call recently about why it is possible for the constraints to hurt performance if the gold answer satisfies the constraints, and we concluded it's because of an interaction with approximate search, but I don't think you've said anything in the paper about what search algorithm or hyperparams you're using.}}
\label{tab:all-lms-non-sql}

\small
\centering
\begin{tabular}{@{}lc c ccc c ccc c ccc}
\toprule
\multirow{2}{*}{\bf LM} & \multirow{2}{1.5cm}{\bf Grammar Constraints} && \multicolumn{3}{c}{\bf SMCalflow} && \multicolumn{3}{c}{\bf TreeDST} && \multicolumn{3}{c}{\bf MTOP (en)}  \\
\cmidrule{4-6}
\cmidrule{8-10}
\cmidrule{12-14}
 &&& Low & Med & High && Low & Med & High && Low & Med & High\\\midrule
 GPT-3$^{\dagger1}$ & Yes && 26.0 & 48.0 & 49.0 && 35.7 & 54.0 & 53.0 && 46.3 & 56.0 & 57.0 \\ 
 Codex$^{\dagger1}$ & Yes && 36.7 & 55.0 & 56.0 && 46.3 & 61.0 & 62.0 && 53.3 & 68.0 & 72.0 \\ 
 Llama-2-7b$^{\dagger1}$ & No && 25.0 & 35.0 & 44.0 && 31.0 & 50.0 & 59.0 && 43.0 & 60.0 & 62.0 \\
 GPT-3$^{\dagger2}$  & Yes && 33.0 & 45.0 & 52.0 && 38.7 & 56.0 & 59.0 && 50.0 & 62.0 & 64.0 \\
 Codex$^{\dagger2}$  & Yes && 39.3 & 52.0 & 62.0 && 47.7 & 58.0 & 64.0 && 55.7 & 67.0 & 70.0 \\
 Llama-2-7b$^{\dagger2}$ & No && 29.0 & 39.0 & 42.0 && 28.0 & 46.0 & 58.0 && 44.0 & 61.0 & 62.0 \\ 
 
\midrule
 \multirow{2}{*}{T5-base} & No && 38.2 & 67.5 & 77.6 && 57.2 & 84.4 & 89.3 && 54.7 & 79.3 & 84.5 \\
 & Yes && 41.6 & 69.7 & 78.6 && 62.0 & 85.8 & 89.4 && 57.5 & 80.1 & 84.3 \\
 \addlinespace
 \multirow{2}{*}{CodeT5-base} & No && 33.3 & 65.6 & 80.8 && 50.3 & 83.5 & 90.3 && 44.1 & 75.6 & 80.6 \\
 & Yes && 37.3 & 67.5 & 81.1 && 56.8 & 84.4 & 90.0 && 47.1 & 75.8 & 81.1 \\
 \addlinespace
 \multirow{2}{*}{BART-large} & No && 36.1 & 68.1 & 82.2 && 52.0 & 84.0 & 90.2 && 57.8 & 81.6 & 85.8 \\
 & Yes && 42.5 & 71.4 & \textbf{83.0} && 61.1 & 86.4 & 89.8 && 61.7 & 82.1 & 85.3 \\
 \addlinespace
 \multirow{2}{*}{T5-large} & No && 42.6 & 71.5 & 81.3 && 59.6 & 86.2 & 90.0 && 55.4 & 82.1 & 85.6 \\
 & Yes && 46.3 & 73.1 & 82.1 && \textbf{64.2} & \textbf{87.2} & 90.1 && 59.3 & 82.5 & 85.3 \\
 \addlinespace
 \multirow{2}{*}{T5-3B} & No && 43.5 & 73.8 & 82.6 && 58.5 & 85.9 & \textbf{90.7} && 60.9 & 83.2 & \textbf{86.3} \\
 & Yes && \textbf{48.7} & \textbf{75.9} & \textbf{83.0} && 64.1 & \textbf{87.2} & 90.3 && \textbf{64.1} & \textbf{83.4} & 85.6 \\\bottomrule

\end{tabular}
\end{table}

\begin{table}[!htbp]
\caption{Performance of fine-tuned language models on Spider, CoSQL and Overnight datasets. We report test suite execution accuracy \cite{ruiqi20} for the SQL datasets and denotation accuracy for Overnight.}
\label{tab:all-lms-sql}

\small
\centering
\begin{tabular}{@{}lc c ccc c ccc c cc@{}}
\toprule
\multirow{2}{*}{\bf LM} & \multirow{2}{1.5cm}{\bf Grammar Constraints?} && \multicolumn{3}{c}{\bf Spider} && \multicolumn{3}{c}{\bf CoSQL} && \multicolumn{2}{c}{\bf Overnight (blocks)}\\
\cmidrule{4-6}
\cmidrule{8-10}
\cmidrule{12-13}
 &&& Low & Med & High && Low & Med & High && Low & High\\\midrule
 \multirow{2}{*}{T5-base} & No && 30.8 & 54.1 & 55.6 && 24.0 & 39.9 & 40.8 && 63.9 & 63.7 \\
 & Yes && 33.8 & 56.8 & 58.9 && 26.8 & 42.7 & 43.4 && \textbf{64.4} & 63.9 \\
 \addlinespace
 \multirow{2}{*}{CodeT5-base} & No && 36.6 & 57.4 & 61.9 && 26.1 & 45.9 & 48.1 && 60.0 & 64.7 \\
 & Yes && 37.6 & 58.0 & 62.2 && 27.3 & 46.2 & 48.2 && 60.4 & 65.2 \\
 \addlinespace
 \multirow{2}{*}{BART-large} & No && 41.8 & 59.1 & 63.9 && 30.9 & 49.9 & 52.8 && 60.7 & 63.4 \\
 & Yes && 42.5 & 62.7 & 63.9 && 29.1 & 48.8 & 51.5 && 61.2 & 63.7 \\
 \addlinespace
 \multirow{2}{*}{T5-large} & No && 42.4 & 64.6 & 65.7 && 30.7 & 50.0 & 53.8 && 62.1 & 68.7 \\
 & Yes && 44.1 & 65.5 & 66.5 && 32.1 & 52.4 & 55.5 && 62.8 & \textbf{68.9} \\
 \addlinespace
 \multirow{2}{*}{T5-3B} & No && 46.4 & 68.4 & 70.9 && 32.6 & 54.7 & 53.4 && 62.8 & 66.2 \\
 & Yes && \textbf{48.6} & \textbf{70.3} & \textbf{72.3} && \textbf{34.7} & \textbf{56.4} & \textbf{56.2} && 63.2 & 66.2 \\\bottomrule
\end{tabular}
\end{table}

\begin{table}[!htbp]
\caption{Performance of fine-tuned language models on constituency (PTB-3), dependency (UD-EWT) and AMR paring. We report bracketing F1 using EVALB \cite{sekine1997evalb} 
for PTB-3, labeled attachment score (LAS) for UD-EWT, and Smatch \cite{cai-knight-2013-smatch} for AMR parsing.}
\label{tab:syntactic-datasets}

\small
\centering
\begin{tabular}{@{}lc c ccc c ccc c ccc@{}}
\toprule
\multirow{2}{*}{\bf LM} & \multirow{2}{1.5cm}{\bf Grammar Constraints?} && \multicolumn{3}{c}{\bf PTB-3} && \multicolumn{3}{c}{\bf UD-EWT} && \multicolumn{3}{c}{\bf AMR 2.0}\\
\cmidrule{4-6}
\cmidrule{8-10}
\cmidrule{12-14}
 &&& Low & Med & High && Low & Med & High && Low & Med & High\\\midrule
 \multirow{2}{*}{T5-base} & No && - & - & - && - & - & - && 52.7 & 72.0 & 75.0 \\
 & Yes && 83.1 & 93.1 & 94.6&& 80.2 & 88.2 & 89.4 && 51.3 & 72.0 & 75.0 \\
 \addlinespace
 \multirow{2}{*}{CodeT5-base} & No && - & - & - && - & - & - && 48.0 & 66.0 & 74.0 \\
 & Yes && 70.9 & 86.7 & 92.1 && 73.2 & 84.0 & 85.8 && 46.7 & 66.0 & 74.0 \\
 \addlinespace
 \multirow{2}{*}{BART-large} & No && - & - & - && - & - & - && 57.0 & 74.0 & 81.0 \\
 & Yes && 78.0 & 93.9 & 95.7 && \textbf{84.1} & 89.7 & 90.6 && 55.0 & 75.0 & 81.0 \\
 \addlinespace
 \multirow{2}{*}{T5-large} & No && - & - & - && - & - & - && 57.3 & 76.0 & 81.0 \\
 & Yes && \textbf{84.5} & \textbf{94.4} & 95.7 && 80.6 & 90.1 & 91.0 && 57.0 & 76.0 & 81.0 \\
 \addlinespace
 \multirow{2}{*}{T5-3B} & No && - & - & - && - & - & - && \textbf{60.0} & \textbf{77.0} & 82.0 \\
 & Yes && 77.6 & 93.9 & \textbf{96.2} && 83.1 & \textbf{90.4} & \textbf{91.3} && 59.0 & \textbf{77.0} & \textbf{83.0} \\\bottomrule
\end{tabular}
\end{table}

\subsection{Format for Model Inputs}
For experiments related to fine-tuned language models with SMCalflow and TreeDST with last user and agent 
utterance as context, the input to the model has the format
$l\text{ }|\text{ }a\text{ }|\text{ }u$, where $u$ is the input natural language
utterance, $l$ is the last user utterance, $a$ is the last agent utterance and $|$ is a separator symbol.
When using only last agent utterance as context, the input is $a\text{ }|\text{ }u$, and for using no context,
the input to the model is simply $u$.

We use the the following format for Spider and CoSQL: $c\text{ },\text{}d\text{ },\text{}u$, where
$c$ is any conversational context if applicable, $d$ is a rendering of the database schema with or without 
values and $u$ is the user utterance. We use the database schema representation used in \citet{scholak-etal-2021-picard} for $d$. For $c$, we concatenate the past utterances in the conversational
context with the separator symbol $|$.

Our few-shot prompting experiments use the prompt template of \citet{shin-etal-2021-constrained}. 
Given a language input, we retrieve relevant prompt examples and create a prompt with the following format:
\vspace{-0.1cm}
\begin{lstlisting}[language=Prompt]
Let's translate what a human user says into what a computer might say.

Human: {Prompt Example 1 Input}
Computer: {Prompt Example 1 Output}
Human: {Prompt Example 2 Input}
Computer: {Prompt Example 2 Output}
              ...
Human: {Current Input}
Computer:
\end{lstlisting}\vspace{1ex}

\subsection{Training Details}
\label{subsec:train-details}
For fine-tuning experiments, we train the language models with batch size \num{32} for \num{10000} steps using AdaFactor \citep{pmlr-v80-shazeer18a}, saving a checkpoint every \num{5000} steps.
We use \num{1000} linear warmup steps and then linear decay the learning rate to 0. We tune all models with learning rates
$10^{-4}$ and $10^{-5}$, except for T5-3B for which we only used $10^{-4}$ to save compute. The best performing checkpoint
on the dev set is used to report scores on the test set.

\section{Results}

\subsection{Benchmarking Language Models}
We show the performance of language models
on BenchCLAMP datasets in tables \ref{tab:all-lms-non-sql}, \ref{tab:all-lms-sql} and \ref{tab:syntactic-datasets}.
We find that performance increases with model size for most fine-tuned language models.
Few-shot prompting of GPT-3 and Codex are still not on par with fine-tuned models.
For non-SQL datasets, even smaller language models reach close to the best performance in
the high resource setting. However, for Spider and CoSQL, model size seems important
in all data settings. This is likely because the model has to reason about the database
schema to generate SQL queries, making it a harder learning problem. See sections \ref{subsec:impact-context} and \ref{subsec:few-shot-promting} for details on how we use context in model inputs for these experiments. We skip unconstrained generation results
for constituency and dependency parsing since models often fail to generate 
valid parses for these tasks, and evaluating a valid substring 
is not supported by released evaluation tools.\samt{I sort of get what you're saying here, but it's unclear. Something like ``Released evaluation tools cannot evaluate invalid outputs.'' tries to say less but is maybe more clear? }
Entirely rejecting invalid 
parses leads to very low performance.

Table \ref{tab:sota} compares our constrained T5-3B model with the best-performing models in the literature. We outperform state-of-the-art models on the SMCalFlow, TreeDST and Overnight-blocks datasets. For Spider and CoSQL, our scores are lower than the state-of-the-art despite using a similar constrained language model approach.
This is likely because we use a general SQL grammar to constrain decoding, whereas the SQL in these datasets covers only a small fraction of the SQL grammar.\jason{but do the Spider and CoSQL models in fact use stronger constraints?} 
We believe using a more constrained grammar will improve performance.%
The state-of-the-art method for AMR also uses a CLAMP model but with additional task specific
pretraining. The best models for PTB-3 and UD-EWT are designed specific to the task. Our language model fine-tuning paired with constrained decoding achieves similar performance without any task specific modifications.

\jason{I think this needs a little more discussion: normally we would like a semantic parser to be able to generalize to new domains, but I guess you are envisioning that someone building a system in a particular domain might only need to use a well-defined subset of SQL.  How would that subset be defined?  Would users still be allowed to ask any SQL query that is valid under the database schema?}

\begin{table}[!htbp]
\caption{Comparison of our fine-tuned T5-3B model with current state of the art models on full test
sets. We report exact match accuracy for Spider and CoSQL to match the settings of previous work. 
The best score in each row is boldfaced.}
\label{tab:sota}
\small
\centering
\begin{tabular}{@{}l p{4cm} r@{}}
\toprule
\bf Dataset & {\bf Current State of the Art} & \bf Our T5-3B \\\midrule
SMCalFlow & 80.4 \cite{platanios-etal-2021-value} & \textbf{83.7} \\
TreeDST & 88.1 \cite{platanios-etal-2021-value} & \textbf{91.5} \\
MTOP (en) & \textbf{86.4} \cite{pasupat-etal-2021-controllable} & 85.7 \\
Overnight & \multirow{2}{*}{65.2 \cite{cao-etal-2019-semantic}} & \multirow{2}{*}{\textbf{66.2}} \\
(blocks) & & \\
Spider & \textbf{75.5} \cite{scholak-etal-2021-picard} & 72.2 \\
CoSQL & \textbf{56.9} \cite{scholak-etal-2021-picard} & 52.3 \\
AMR & \textbf{85.4} \cite{bai-etal-2022-graph} & 83.0 \\ \midrule
PTB 3 & \textbf{96.4} \cite{tian-etal-2020-improving} & 96.2 \\
UD-EWT & \textbf{91.5} \cite{mohammadshahi-henderson-2021-recursive} & 91.3 \\\bottomrule
\end{tabular}
\end{table}

\subsection{Effect of Constraints} 
\label{subsec:effect-constraints}
Decoding constrained by a grammar is essential for few-shot prompted models
like GPT-3 and Codex that are accessed via API calls. Without a grammar, we noticed that these models explore a large number of invalid paths leading to 
high latency and API cost. We also found constrained decoding essential for
source side prediction tasks like constituency and dependency parsing. They require each token
in the input to be covered in the output. We found even fine-tuned language models
struggle to learn these constraints leading to very low performance with
unconstrained decoding.

Tables \ref{tab:all-lms-non-sql}, \ref{tab:all-lms-sql} and \ref{tab:syntactic-datasets} show the
effect of constraints while generating from fine-tuned large language models. We find that constrained decoding is most beneficial in low-data regimes, giving on average $2.7\%$
gain over unconstrained decoding. In the high-resource setting, the average gain is less
than $1\%$, suggesting that the full data is nearly sufficient to learn the constraint system. 

We find that constrained decoding under performs
unconstrained decoding for some settings. We attribute
this to the insufficient coverage of our
grammars. Our grammars were induced from the training 
data, and thereby fail to cover novel components
or combinations at test time. They are also 
constrained to copy quoted strings from the input utterance. However this is not strictly followed
in some datasets. Table \ref{tab:grammar-coverage}
shows the fraction of test outputs covered by our
grammars. We could relax the grammar constraints
to ensure full coverage; we leave such exploration to
future work. In 
realistic situations, we expect grammars to be 
provided by domain developers ensuring
full coverage.
\begin{table}[!htbp]
\caption{Grammar coverage (\%) of the gold outputs in the test set of
BenchCLAMP datasets.}
\label{tab:grammar-coverage}
\small
\centering
\begin{tabular}{@{}l r@{}}
\toprule
\bf Dataset & \bf Test Set Coverage \%\\\midrule
SMCalFlow & 99.6\\
TreeDST & 96.3 \\
MTOP (en) & 96.6 \\
Overnight (blocks) & 100.0 \\
Spider & 98.8 \\
CoSQL & 99.4 \\
AMR & 99.7 \\ \midrule
PTB 3 & 100.0 \\
UD-EWT & 99.9 \\\bottomrule
\end{tabular}
\end{table}


\subsection{Impact of Context} 
\label{subsec:impact-context}
The datasets in BenchCLAMP require a model to use a variety of
contexts.
SMCalFlow, TreeDST and CoSQL datasets all have conversational
context. Spider and CoSQL have database schema context which informs the target SQL 
prediction.  BenchCLAMP allows us to perform a controlled investigation of the effect of context.
Table \ref{tab:lispress-context} shows
that while using the last agent and user utterance is helpful for all settings, the low-data regime does even better when using only the last agent utterance; without more data, training struggles to 
learn how to utilize (or ignore) the additional context.
We find similar results for CoSQL in Table \ref{tab:cosql-context}.
Also, SQL prediction always benefits from including database values in the context along with the database schema information. 
We use the best settings for context for each data regime for benchmarking results
in tables \ref{tab:all-lms-non-sql}, \ref{tab:all-lms-sql} and \ref{tab:syntactic-datasets}.

\begin{table}[!htbp]
\caption{Lispress match accuracy of unconstrained fine-tuned T5-large with varying conversational context on SMCalFlow and TreeDST. We find more context hurts in low resource settings but helps in medium and high resource settings. 
\adpauls{If you have the space, it would be really nice to add a sentence about the main conclusion (hurts for low-resource, helps for mid and high).}
}
\label{tab:lispress-context}
\small
\centering
\begin{tabular}{@{}l ccc ccc@{}}
\toprule
\multirow{2}{*}{\bf Conv. Context} & \multicolumn{3}{c}{\bf SMCalFlow} & \multicolumn{3}{c}{\bf TreeDST} \\
\cmidrule(lr){2-4} \cmidrule(lr){5-7}
 & Low & Med & High & Low & Med & High\\
 \midrule
 No context & 37.0 & 63.8 & 72.9 & 42.4 & 68.4 & 76.0 \\
 Last agent utt. & \textbf{42.6} & 70.7 & 80.6 & \textbf{59.6} & 82.7 & 87.9 \\
 Last user \& agent utt. & 40.0 & \textbf{71.5} & \textbf{81.3} & 58.8 & \textbf{86.2} & \textbf{90.0} \\\bottomrule
\end{tabular}
\end{table}

\begin{table}[!htbp]
\caption{Test suite execution accuracy of unconstrained BART-large  
on CoSQL with varying context. More context hurts
in low resource settings but helps in medium and high
resource settings.}
\label{tab:cosql-context}
\small
\centering
\begin{tabular}{@{}p{2cm} p{1.5cm} ccc@{}}
\toprule
\multirow{2}{2cm}{\bf Conversational Context} & \multirow{2}{*}{\bf DB values?} & \multirow{2}{*}{\bf Low} & \multirow{2}{*}{\bf Med} & \multirow{2}{*}{\bf High} \\
& & & & \\
\midrule
 \multirow{2}{*}{No context} & no & 21.3 & 35.9 & 34.4 \\
 & yes & 25.3 & 38.9 & 40.3 \\
\addlinespace

 \multirow{2}{*}{Last interaction} & no & 24.1 & 40.4 & 39.1 \\
 & yes & \textbf{28.2} & 44.2 & 44.4 \\
\addlinespace

 \multirow{2}{*}{All interactions} & no & 24.3 & 36.8 & 43.0 \\
 & yes & 24.9 & \textbf{44.9} & \textbf{48.8} \\\bottomrule
\end{tabular}
\end{table}

\subsection{Few-Shot Prompting} 
\label{subsec:few-shot-promting}
In few-shot prompting scenario, we manipulate the context choice and ordering of examples in our prompt to Codex. The results in Table \ref{tab:gpt3-ablations} show that ordering the most similar example at the end closest
to the generation heads is helpful in the low-data regime, indicating that GPT-3 and Codex pay more attention to the recent past. In higher data regime, all prompt examples are almost equally relevant, hence the order does not matter as much. We find that context does not help; one of the reasons being that we can fit fewer examples in the prompt if we include context for each example. 

We experiment with BM25 and similarity models for prompt retrieval for few-shot prompting. For similarity models, we pick top models from each of the three
categories in SentenceTransformers leaderboard \cite{reimers-2019-sentence-bert}:
all-mpnet-base-v2, multi-qa-mpnet-base-dot-v1 and sentence-t5-xxl \cite{ni-etal-2022-sentence-t5}. We find that SentenceT5-xxl surpasses other similarity models
for prompt retrieval. SentenceT5 outperforms BM25 in low resource settings, but 
performs relatively worse when more data is available.

\begin{table}[!htbp]
\caption{(Left): Lispress match accuracy of the few-shot prompted
Codex DaVinci model on SMCalFlow with different 
prompt order and conversational context.
More context hurts few-shot prompted models. Ordering the most relevant 
examples closer to the generation heads improve performance.
(Right): Lispress match accuracy of Codex on SMCalFlow with different prompt retrieval methods. We used \texttt{code-davinci-001} with best last prompt order and
no context.}
\label{tab:gpt3-ablations}
\small
\centering
\begin{tabular}{@{}p{1.9cm} p{2.5cm} cc@{}}
\toprule
\bf Prompt Order & \bf Conv. Context & \bf Low & \bf Med \\\midrule
Random & No context & 35.7 & 52.0 \\
Best First & No context & 34.3 & \textbf{53.0} \\
Best Last & No context & \textbf{36.7} & 52.0 \\
Best Last & Last agent utt. & 34.0 & 41.0 \\
Best Last & Last user \& agent utt. & 26.0 & 31.0 \\\bottomrule
\end{tabular}
\quad\quad
\begin{tabular}{p{2.5cm}cc}
\toprule
\bf Prompt Retrieval Method & \bf Low & \bf Med \\\midrule
BM25 & 36.7 & \textbf{55.0} \\
all-mpnet-base-v2 & 38.0 & 49.0 \\
multi-qa-mpnet-base-dot-v1 & 38.3 & 50.0 \\
sentence-t5-xxl & \textbf{39.3} & 52.0 \\\bottomrule
\end{tabular}
\end{table}

\begin{table}[!htbp]
\caption{Effect of different grammar induction data on the Lispress match constrained decoding accuracy of fine-tuned T5-large.}
\label{tab:small-grammar}
\small
\centering
\begin{tabular}{@{}p{3cm} cc cc@{}}
\toprule
\multirow{2}{3cm}{\bf Constraint Grammar} & \multicolumn{2}{c }{\bf SMCalFlow} & \multicolumn{2}{c}{\bf TreeDST} \\
\cmidrule(lr){2-3} \cmidrule(lr){4-5}
 & Low & Med & Low & Med \\\midrule
 Unconstrained & 42.6 & 71.5 & 59.6 & 86.2 \\
 Induced from train split & 45.6 & 73.1 & 62.3 & 87.2 \\
 Induced from full train & 46.3 & 73.1 & 64.2 & 87.2 \\\bottomrule
\end{tabular}
\end{table}

\subsection{Grammars induced from Less Data}
\label{subsec:low-resource-grammar}
The grammars for SMCalFlow, TreeDST and MTOP were induced using the full train dataset. This grammar
is then used even with low and medium resource train splits. We expect the grammar will be provided by a developer of 
the domain and hence will cover all valid representations. For completeness, we
report here the impact of using grammar induced from the corresponding train sets.
Table \ref{tab:small-grammar} shows the results of decoding with train split induced grammar, and
compares the performance with unconstrained decoding and decoding with grammar induced from full train set. 
The gains from constraints drop by $1-2\%$ for low resource splits when using train split induced grammar
instead of full train induced grammar. It does not affect results for medium resource splits.

\section{Conclusion}
We introduce a benchmark comprising nine parsing datasets with varying target representations. 
We support few-shot prompting, fine-tuning and constrained decoding for all autoregressive language models and sequence-to-sequence models on these datasets. We hope that
this work will encourage language model developers to consider parsing tasks as a test-bed
in future work.

\section{Limitations}
\label{sec:limitations}
Our benchmark includes data in multiple languages (all languages included
in the MTOP dataset) but we only evaluate on English datasets due to compute
constraints.
Few-shot prompted experiments were evaluated on relatively small test
sets on three datasets due to API cost limitations. As a result, we noticed high variance in the 
results (see Appendix section \ref{sec:variance} for variance results). 

\section*{Acknowledgements}
Thanks to Yunmo Chen who helped set up Llama-2 models for our benchmark experiments.

\bibliographystyle{plainnat}
\bibliography{references}

\begin{thebibliography}{54}
\providecommand{\natexlab}[1]{#1}
\providecommand{\url}[1]{\texttt{#1}}
\expandafter\ifx\csname urlstyle\endcsname\relax
  \providecommand{\doi}[1]{doi: #1}\else
  \providecommand{\doi}{doi: \begingroup \urlstyle{rm}\Url}\fi

\bibitem[Andreas et~al.(2020)Andreas, Bufe, Burkett, Chen, Clausman, Crawford,
  Crim, DeLoach, Dorner, Eisner, Fang, Guo, Hall, Hayes, Hill, Ho, Iwaszuk,
  Jha, Klein, Krishnamurthy, Lanman, Liang, Lin, Lintsbakh, McGovern,
  Nisnevich, Pauls, Petters, Read, Roth, Roy, Rusak, Short, Slomin, Snyder,
  Striplin, Su, Tellman, Thomson, Vorobev, Witoszko, Wolfe, Wray, Zhang, and
  Zotov]{andreas-etal-2020-task}
Jacob Andreas, John Bufe, David Burkett, Charles Chen, Josh Clausman, Jean
  Crawford, Kate Crim, Jordan DeLoach, Leah Dorner, Jason Eisner, Hao Fang,
  Alan Guo, David Hall, Kristin Hayes, Kellie Hill, Diana Ho, Wendy Iwaszuk,
  Smriti Jha, Dan Klein, Jayant Krishnamurthy, Theo Lanman, Percy Liang,
  Christopher~H. Lin, Ilya Lintsbakh, Andy McGovern, Aleksandr Nisnevich, Adam
  Pauls, Dmitrij Petters, Brent Read, Dan Roth, Subhro Roy, Jesse Rusak, Beth
  Short, Div Slomin, Ben Snyder, Stephon Striplin, Yu~Su, Zachary Tellman, Sam
  Thomson, Andrei Vorobev, Izabela Witoszko, Jason Wolfe, Abby Wray, Yuchen
  Zhang, and Alexander Zotov.
\newblock Task-oriented dialogue as dataflow synthesis.
\newblock \emph{Transactions of the Association for Computational Linguistics},
  8:\penalty0 556--571, 2020.
\newblock \doi{10.1162/tacl_a_00333}.
\newblock URL \url{https://www.aclweb.org/anthology/2020.tacl-1.36}.

\bibitem[antlr(2022)]{antlr-2022}
antlr.
\newblock grammars-v4.
\newblock \url{https://github.com/antlr/grammars-v4}, 2022.

\bibitem[Bai et~al.(2022)Bai, Chen, and Zhang]{bai-etal-2022-graph}
Xuefeng Bai, Yulong Chen, and Yue Zhang.
\newblock Graph pre-training for {AMR} parsing and generation.
\newblock In \emph{Proceedings of the 60th Annual Meeting of the Association
  for Computational Linguistics (Volume 1: Long Papers)}, pages 6001--6015,
  Dublin, Ireland, May 2022. Association for Computational Linguistics.
\newblock \doi{10.18653/v1/2022.acl-long.415}.
\newblock URL \url{https://aclanthology.org/2022.acl-long.415}.

\bibitem[Banarescu et~al.(2013)Banarescu, Bonial, Cai, Georgescu, Griffitt,
  Hermjakob, Knight, Koehn, Palmer, and Schneider]{banarescu-etal-2013-AMR}
Laura Banarescu, Claire Bonial, Shu Cai, Madalina Georgescu, Kira Griffitt, Ulf
  Hermjakob, Kevin Knight, Philipp Koehn, Martha Palmer, and Nathan Schneider.
\newblock {A}bstract {M}eaning {R}epresentation for sembanking.
\newblock In \emph{Proceedings of the 7th Linguistic Annotation Workshop and
  Interoperability with Discourse}, pages 178--186, Sofia, Bulgaria, August
  2013. Association for Computational Linguistics.
\newblock URL \url{https://aclanthology.org/W13-2322}.

\bibitem[Brown et~al.(2020)Brown, Mann, Ryder, Subbiah, Kaplan, Dhariwal,
  Neelakantan, Shyam, Sastry, Askell, Agarwal, Herbert-Voss, Krueger, Henighan,
  Child, Ramesh, Ziegler, Wu, Winter, Hesse, Chen, Sigler, Litwin, Gray, Chess,
  Clark, Berner, McCandlish, Radford, Sutskever, and Amodei]{Brown:2020:gpt-3}
Tom~B. Brown, Benjamin Mann, Nick Ryder, Melanie Subbiah, Jared Kaplan,
  Prafulla Dhariwal, Arvind Neelakantan, Pranav Shyam, Girish Sastry, Amanda
  Askell, Sandhini Agarwal, Ariel Herbert-Voss, Gretchen Krueger, Tom Henighan,
  Rewon Child, Aditya Ramesh, Daniel~M. Ziegler, Jeffrey Wu, Clemens Winter,
  Christopher Hesse, Mark Chen, Eric Sigler, Mateusz Litwin, Scott Gray,
  Benjamin Chess, Jack Clark, Christopher Berner, Sam McCandlish, Alec Radford,
  Ilya Sutskever, and Dario Amodei.
\newblock Language models are few-shot learners.
\newblock \emph{Computing Research Repository}, arXiv:2005.14165, 2020.

\bibitem[Cai and Lam(2020)]{cai-lam-2020-amr}
Deng Cai and Wai Lam.
\newblock {AMR} parsing via graph-sequence iterative inference.
\newblock In \emph{Proceedings of the 58th Annual Meeting of the Association
  for Computational Linguistics}, pages 1290--1301, Online, July 2020.
  Association for Computational Linguistics.
\newblock \doi{10.18653/v1/2020.acl-main.119}.
\newblock URL \url{https://aclanthology.org/2020.acl-main.119}.

\bibitem[Cai and Knight(2013)]{cai-knight-2013-smatch}
Shu Cai and Kevin Knight.
\newblock {S}match: an evaluation metric for semantic feature structures.
\newblock In \emph{Proceedings of the 51st Annual Meeting of the Association
  for Computational Linguistics (Volume 2: Short Papers)}, pages 748--752,
  Sofia, Bulgaria, August 2013. Association for Computational Linguistics.
\newblock URL \url{https://www.aclweb.org/anthology/P13-2131}.

\bibitem[Cao et~al.(2019)Cao, Zhu, Liu, Li, and Yu]{cao-etal-2019-semantic}
Ruisheng Cao, Su~Zhu, Chen Liu, Jieyu Li, and Kai Yu.
\newblock Semantic parsing with dual learning.
\newblock In \emph{Proceedings of the 57th Annual Meeting of the Association
  for Computational Linguistics}, pages 51--64, Florence, Italy, July 2019.
  Association for Computational Linguistics.
\newblock \doi{10.18653/v1/P19-1007}.
\newblock URL \url{https://www.aclweb.org/anthology/P19-1007}.

\bibitem[Chen et~al.(2021)Chen, Tworek, Jun, Yuan, de~Oliveira~Pinto, Kaplan,
  Edwards, Burda, Joseph, Brockman, Ray, Puri, Krueger, Petrov, Khlaaf, Sastry,
  Mishkin, Chan, Gray, Ryder, Pavlov, Power, Kaiser, Bavarian, Winter, Tillet,
  Such, Cummings, Plappert, Chantzis, Barnes, Herbert{-}Voss, Guss, Nichol,
  Paino, Tezak, Tang, Babuschkin, Balaji, Jain, Saunders, Hesse, Carr, Leike,
  Achiam, Misra, Morikawa, Radford, Knight, Brundage, Murati, Mayer, Welinder,
  McGrew, Amodei, McCandlish, Sutskever, and Zaremba]{codex-chen-2021}
Mark Chen, Jerry Tworek, Heewoo Jun, Qiming Yuan, Henrique~Ponde
  de~Oliveira~Pinto, Jared Kaplan, Harrison Edwards, Yuri Burda, Nicholas
  Joseph, Greg Brockman, Alex Ray, Raul Puri, Gretchen Krueger, Michael Petrov,
  Heidy Khlaaf, Girish Sastry, Pamela Mishkin, Brooke Chan, Scott Gray, Nick
  Ryder, Mikhail Pavlov, Alethea Power, Lukasz Kaiser, Mohammad Bavarian,
  Clemens Winter, Philippe Tillet, Felipe~Petroski Such, Dave Cummings,
  Matthias Plappert, Fotios Chantzis, Elizabeth Barnes, Ariel Herbert{-}Voss,
  William~Hebgen Guss, Alex Nichol, Alex Paino, Nikolas Tezak, Jie Tang, Igor
  Babuschkin, Suchir Balaji, Shantanu Jain, William Saunders, Christopher
  Hesse, Andrew~N. Carr, Jan Leike, Joshua Achiam, Vedant Misra, Evan Morikawa,
  Alec Radford, Matthew Knight, Miles Brundage, Mira Murati, Katie Mayer, Peter
  Welinder, Bob McGrew, Dario Amodei, Sam McCandlish, Ilya Sutskever, and
  Wojciech Zaremba.
\newblock Evaluating large language models trained on code.
\newblock \emph{CoRR}, abs/2107.03374, 2021.
\newblock URL \url{https://arxiv.org/abs/2107.03374}.

\bibitem[Cheng et~al.(2020)Cheng, Agrawal, Mart{\'\i}nez~Alonso, Bhargava,
  Driesen, Flego, Kaplan, Kartsaklis, Li, Piraviperumal, Williams, Yu,
  {\'O}~S{\'e}aghdha, and Johannsen]{cheng-etal-2020-conversational}
Jianpeng Cheng, Devang Agrawal, H{\'e}ctor Mart{\'\i}nez~Alonso, Shruti
  Bhargava, Joris Driesen, Federico Flego, Dain Kaplan, Dimitri Kartsaklis, Lin
  Li, Dhivya Piraviperumal, Jason~D. Williams, Hong Yu, Diarmuid
  {\'O}~S{\'e}aghdha, and Anders Johannsen.
\newblock Conversational semantic parsing for dialog state tracking.
\newblock In \emph{Proceedings of the 2020 Conference on Empirical Methods in
  Natural Language Processing (EMNLP)}, pages 8107--8117, Online, November
  2020. Association for Computational Linguistics.
\newblock \doi{10.18653/v1/2020.emnlp-main.651}.
\newblock URL \url{https://aclanthology.org/2020.emnlp-main.651}.

\bibitem[Earley(1970)]{earley70}
Jay Earley.
\newblock An efficient context-free parsing algorithm.
\newblock \emph{Communications of the ACM}, 13\penalty0 (2):\penalty0 94--102,
  1970.
\newblock \doi{10.1145/362007.362035}.
\newblock URL \url{https://doi.org/10.1145/362007.362035}.

\bibitem[Fern{\'a}ndez-Gonz{\'a}lez and
  G{\'o}mez-Rodr{\'\i}guez(2020)]{fernandez-gonzalez-gomez-rodriguez-2020-enriched}
Daniel Fern{\'a}ndez-Gonz{\'a}lez and Carlos G{\'o}mez-Rodr{\'\i}guez.
\newblock Enriched in-order linearization for faster sequence-to-sequence
  constituent parsing.
\newblock In \emph{Proceedings of the 58th Annual Meeting of the Association
  for Computational Linguistics}, pages 4092--4099, Online, July 2020.
  Association for Computational Linguistics.
\newblock \doi{10.18653/v1/2020.acl-main.376}.
\newblock URL \url{https://aclanthology.org/2020.acl-main.376}.

\bibitem[He et~al.(2021)He, Gao, and Chen]{deberta-v3-he-2021}
Pengcheng He, Jianfeng Gao, and Weizhu Chen.
\newblock Debertav3: Improving deberta using electra-style pre-training with
  gradient-disentangled embedding sharing.
\newblock \emph{CoRR}, abs/2111.09543, 2021.
\newblock URL \url{https://arxiv.org/abs/2111.09543}.

\bibitem[Kitaev and Klein(2018)]{kitaev-klein-2018-constituency}
Nikita Kitaev and Dan Klein.
\newblock Constituency parsing with a self-attentive encoder.
\newblock In \emph{Proceedings of the 56th Annual Meeting of the Association
  for Computational Linguistics (Volume 1: Long Papers)}, pages 2676--2686,
  Melbourne, Australia, July 2018. Association for Computational Linguistics.
\newblock \doi{10.18653/v1/P18-1249}.
\newblock URL \url{https://aclanthology.org/P18-1249}.

\bibitem[Koo et~al.(2015)Koo, Petrov, Sutskever, Hinton, Vinyals, and
  Kaiser]{koo2015grammar}
Petrov Koo, S~Petrov, I~Sutskever, GE~Hinton, O~Vinyals, and L~Kaiser.
\newblock Grammar as a foreign language.
\newblock In \emph{Advances in Neural Information Processing Systems}, 2015.

\bibitem[Kwon et~al.(2023)Kwon, Li, Zhuang, Sheng, Zheng, Yu, Gonzalez, Zhang,
  and Stoica]{vllm}
Woosuk Kwon, Zhuohan Li, Siyuan Zhuang, Ying Sheng, Lianmin Zheng, Cody~Hao Yu,
  Joseph Gonzalez, Hao Zhang, and Ion Stoica.
\newblock Efficient memory management for large language model serving with
  pagedattention.
\newblock In Jason Flinn, Margo~I. Seltzer, Peter Druschel, Antoine Kaufmann,
  and Jonathan Mace, editors, \emph{Proceedings of the 29th Symposium on
  Operating Systems Principles, {SOSP} 2023, Koblenz, Germany, October 23-26,
  2023}, pages 611--626. {ACM}, 2023.
\newblock \doi{10.1145/3600006.3613165}.
\newblock URL \url{https://doi.org/10.1145/3600006.3613165}.

\bibitem[Lewis et~al.(2020)Lewis, Liu, Goyal, Ghazvininejad, Mohamed, Levy,
  Stoyanov, and Zettlemoyer]{lewis-etal-2020-bart}
Mike Lewis, Yinhan Liu, Naman Goyal, Marjan Ghazvininejad, Abdelrahman Mohamed,
  Omer Levy, Veselin Stoyanov, and Luke Zettlemoyer.
\newblock {BART}: Denoising sequence-to-sequence pre-training for natural
  language generation, translation, and comprehension.
\newblock In \emph{Proceedings of the 58th Annual Meeting of the Association
  for Computational Linguistics}, pages 7871--7880, Online, July 2020.
  Association for Computational Linguistics.
\newblock \doi{10.18653/v1/2020.acl-main.703}.
\newblock URL \url{https://www.aclweb.org/anthology/2020.acl-main.703}.

\bibitem[Li et~al.(2021)Li, Arora, Chen, Gupta, Gupta, and
  Mehdad]{mtop-li-2021}
Haoran Li, Abhinav Arora, Shuohui Chen, Anchit Gupta, Sonal Gupta, and Yashar
  Mehdad.
\newblock {MTOP}: A comprehensive multilingual task-oriented semantic parsing
  benchmark.
\newblock In \emph{Proceedings of the 16th Conference of the European Chapter
  of the Association for Computational Linguistics: Main Volume}, pages
  2950--2962, Online, April 2021. Association for Computational Linguistics.
\newblock \doi{10.18653/v1/2021.eacl-main.257}.
\newblock URL \url{https://aclanthology.org/2021.eacl-main.257}.

\bibitem[Li et~al.(2018)Li, Cai, He, and Zhao]{li-etal-2018-seq2seq}
Zuchao Li, Jiaxun Cai, Shexia He, and Hai Zhao.
\newblock Seq2seq dependency parsing.
\newblock In \emph{Proceedings of the 27th International Conference on
  Computational Linguistics}, pages 3203--3214, Santa Fe, New Mexico, USA,
  August 2018. Association for Computational Linguistics.
\newblock URL \url{https://aclanthology.org/C18-1271}.

\bibitem[Liang et~al.(2022)Liang, Bommasani, Lee, Tsipras, Soylu, Yasunaga,
  Zhang, Narayanan, Wu, Kumar, Newman, Yuan, Yan, Zhang, Cosgrove, Manning,
  Ré, Acosta-Navas, Hudson, Zelikman, Durmus, Ladhak, Rong, Ren, Yao, Wang,
  Santhanam, Orr, Zheng, Yuksekgonul, Suzgun, Kim, Guha, Chatterji, Khattab,
  Henderson, Huang, Chi, Xie, Santurkar, Ganguli, Hashimoto, Icard, Zhang,
  Chaudhary, Wang, Li, Mai, Zhang, and Koreeda]{helm}
Percy Liang, Rishi Bommasani, Tony Lee, Dimitris Tsipras, Dilara Soylu,
  Michihiro Yasunaga, Yian Zhang, Deepak Narayanan, Yuhuai Wu, Ananya Kumar,
  Benjamin Newman, Binhang Yuan, Bobby Yan, Ce~Zhang, Christian Cosgrove,
  Christopher~D. Manning, Christopher Ré, Diana Acosta-Navas, Drew~A. Hudson,
  Eric Zelikman, Esin Durmus, Faisal Ladhak, Frieda Rong, Hongyu Ren, Huaxiu
  Yao, Jue Wang, Keshav Santhanam, Laurel Orr, Lucia Zheng, Mert Yuksekgonul,
  Mirac Suzgun, Nathan Kim, Neel Guha, Niladri Chatterji, Omar Khattab, Peter
  Henderson, Qian Huang, Ryan Chi, Sang~Michael Xie, Shibani Santurkar, Surya
  Ganguli, Tatsunori Hashimoto, Thomas Icard, Tianyi Zhang, Vishrav Chaudhary,
  William Wang, Xuechen Li, Yifan Mai, Yuhui Zhang, and Yuta Koreeda.
\newblock Holistic evaluation of language models, 2022.
\newblock URL \url{https://arxiv.org/abs/2211.09110}.

\bibitem[Liu and Zhang(2017)]{liu-zhang-transition-2017}
Jiangming Liu and Yue Zhang.
\newblock {In-Order Transition-based Constituent Parsing}.
\newblock \emph{Transactions of the Association for Computational Linguistics},
  5:\penalty0 413--424, 11 2017.
\newblock ISSN 2307-387X.
\newblock \doi{10.1162/tacl_a_00070}.
\newblock URL \url{https://doi.org/10.1162/tacl\_a\_00070}.

\bibitem[Liu et~al.(2019)Liu, Ott, Goyal, Du, Joshi, Chen, Levy, Lewis,
  Zettlemoyer, and Stoyanov]{roberta-liu-2019}
Yinhan Liu, Myle Ott, Naman Goyal, Jingfei Du, Mandar Joshi, Danqi Chen, Omer
  Levy, Mike Lewis, Luke Zettlemoyer, and Veselin Stoyanov.
\newblock Roberta: A robustly optimized bert pretraining approach.
\newblock \emph{arXiv preprint arXiv:1907.11692}, 2019.

\bibitem[Marcus et~al.(1993)Marcus, Santorini, and
  Marcinkiewicz]{marcus-etal-1993-ptb}
Mitchell~P. Marcus, Beatrice Santorini, and Mary~Ann Marcinkiewicz.
\newblock Building a large annotated corpus of {E}nglish: The {P}enn
  {T}reebank.
\newblock \emph{Computational Linguistics}, 19\penalty0 (2):\penalty0 313--330,
  1993.
\newblock URL \url{https://aclanthology.org/J93-2004}.

\bibitem[Mohammadshahi and
  Henderson(2021)]{mohammadshahi-henderson-2021-recursive}
Alireza Mohammadshahi and James Henderson.
\newblock Recursive non-autoregressive graph-to-graph transformer for
  dependency parsing with iterative refinement.
\newblock \emph{Transactions of the Association for Computational Linguistics},
  9:\penalty0 120--138, 2021.
\newblock \doi{10.1162/tacl_a_00358}.
\newblock URL \url{https://aclanthology.org/2021.tacl-1.8}.

\bibitem[Ni et~al.(2022)Ni, Hernandez~Abrego, Constant, Ma, Hall, Cer, and
  Yang]{ni-etal-2022-sentence-t5}
Jianmo Ni, Gustavo Hernandez~Abrego, Noah Constant, Ji~Ma, Keith Hall, Daniel
  Cer, and Yinfei Yang.
\newblock Sentence-t5: Scalable sentence encoders from pre-trained text-to-text
  models.
\newblock In \emph{Findings of the Association for Computational Linguistics:
  ACL 2022}, pages 1864--1874, Dublin, Ireland, May 2022. Association for
  Computational Linguistics.
\newblock \doi{10.18653/v1/2022.findings-acl.146}.
\newblock URL \url{https://aclanthology.org/2022.findings-acl.146}.

\bibitem[Pasupat et~al.(2021)Pasupat, Zhang, and
  Guu]{pasupat-etal-2021-controllable}
Panupong Pasupat, Yuan Zhang, and Kelvin Guu.
\newblock Controllable semantic parsing via retrieval augmentation.
\newblock In \emph{Proceedings of the 2021 Conference on Empirical Methods in
  Natural Language Processing}, pages 7683--7698, Online and Punta Cana,
  Dominican Republic, November 2021. Association for Computational Linguistics.
\newblock \doi{10.18653/v1/2021.emnlp-main.607}.
\newblock URL \url{https://aclanthology.org/2021.emnlp-main.607}.

\bibitem[Platanios et~al.(2021)Platanios, Pauls, Roy, Zhang, Kyte, Guo,
  Thomson, Krishnamurthy, Wolfe, Andreas, and Klein]{platanios-etal-2021-value}
Emmanouil~Antonios Platanios, Adam Pauls, Subhro Roy, Yuchen Zhang, Alexander
  Kyte, Alan Guo, Sam Thomson, Jayant Krishnamurthy, Jason Wolfe, Jacob
  Andreas, and Dan Klein.
\newblock Value-agnostic conversational semantic parsing.
\newblock In \emph{Proceedings of the 59th Annual Meeting of the Association
  for Computational Linguistics and the 11th International Joint Conference on
  Natural Language Processing (Volume 1: Long Papers)}, pages 3666--3681,
  Online, August 2021. Association for Computational Linguistics.
\newblock \doi{10.18653/v1/2021.acl-long.284}.
\newblock URL \url{https://aclanthology.org/2021.acl-long.284}.

\bibitem[Raffel et~al.(2020)Raffel, Shazeer, Roberts, Lee, Narang, Matena,
  Zhou, Li, and Liu]{t5-raffel-2020}
Colin Raffel, Noam Shazeer, Adam Roberts, Katherine Lee, Sharan Narang, Michael
  Matena, Yanqi Zhou, Wei Li, and Peter~J. Liu.
\newblock Exploring the limits of transfer learning with a unified text-to-text
  transformer.
\newblock \emph{Journal of Machine Learning Research}, 21\penalty0
  (140):\penalty0 1--67, 2020.
\newblock URL \url{http://jmlr.org/papers/v21/20-074.html}.

\bibitem[Reimers and Gurevych(2019)]{reimers-2019-sentence-bert}
Nils Reimers and Iryna Gurevych.
\newblock Sentence-bert: Sentence embeddings using siamese bert-networks.
\newblock In \emph{Proceedings of the 2019 Conference on Empirical Methods in
  Natural Language Processing}. Association for Computational Linguistics, 11
  2019.
\newblock URL \url{https://arxiv.org/abs/1908.10084}.

\bibitem[Rubin et~al.(2021)Rubin, Herzig, and
  Berant]{rubin-2021-in-context-learning}
Ohad Rubin, Jonathan Herzig, and Jonathan Berant.
\newblock Learning to retrieve prompts for in-context learning.
\newblock \emph{CoRR}, abs/2112.08633, 2021.
\newblock URL \url{https://arxiv.org/abs/2112.08633}.

\bibitem[Sanh et~al.(2021)Sanh, Webson, Raffel, Bach, Sutawika, Alyafeai,
  Chaffin, Stiegler, Scao, Raja, Dey, Bari, Xu, Thakker, Sharma, Szczechla,
  Kim, Chhablani, Nayak, Datta, Chang, Jiang, Wang, Manica, Shen, Yong, Pandey,
  Bawden, Wang, Neeraj, Rozen, Sharma, Santilli, Fevry, Fries, Teehan,
  Biderman, Gao, Bers, Wolf, and Rush]{sanh2021multitask}
Victor Sanh, Albert Webson, Colin Raffel, Stephen~H. Bach, Lintang Sutawika,
  Zaid Alyafeai, Antoine Chaffin, Arnaud Stiegler, Teven~Le Scao, Arun Raja,
  Manan Dey, M~Saiful Bari, Canwen Xu, Urmish Thakker, Shanya~Sharma Sharma,
  Eliza Szczechla, Taewoon Kim, Gunjan Chhablani, Nihal Nayak, Debajyoti Datta,
  Jonathan Chang, Mike Tian-Jian Jiang, Han Wang, Matteo Manica, Sheng Shen,
  Zheng~Xin Yong, Harshit Pandey, Rachel Bawden, Thomas Wang, Trishala Neeraj,
  Jos Rozen, Abheesht Sharma, Andrea Santilli, Thibault Fevry, Jason~Alan
  Fries, Ryan Teehan, Stella Biderman, Leo Gao, Tali Bers, Thomas Wolf, and
  Alexander~M. Rush.
\newblock Multitask prompted training enables zero-shot task generalization,
  2021.

\bibitem[Scholak et~al.(2021)Scholak, Schucher, and
  Bahdanau]{scholak-etal-2021-picard}
Torsten Scholak, Nathan Schucher, and Dzmitry Bahdanau.
\newblock {PICARD}: Parsing incrementally for constrained auto-regressive
  decoding from language models.
\newblock In \emph{Proceedings of the 2021 Conference on Empirical Methods in
  Natural Language Processing}, pages 9895--9901, Online and Punta Cana,
  Dominican Republic, November 2021. Association for Computational Linguistics.
\newblock \doi{10.18653/v1/2021.emnlp-main.779}.
\newblock URL \url{https://aclanthology.org/2021.emnlp-main.779}.

\bibitem[Schucher et~al.(2021)Schucher, Reddy, and de~Vries]{schucher-2021}
Nathan Schucher, Siva Reddy, and Harm de~Vries.
\newblock The power of prompt tuning for low-resource semantic parsing.
\newblock \emph{CoRR}, abs/2110.08525, 2021.
\newblock URL \url{https://arxiv.org/abs/2110.08525}.

\bibitem[Sekine and Collins(1997)]{sekine1997evalb}
Satoshi Sekine and Michael Collins.
\newblock Evalb bracket scoring program.
\newblock \emph{URL: http://www. cs. nyu. edu/cs/projects/proteus/evalb}, 1997.

\bibitem[Shazeer and Stern(2018)]{pmlr-v80-shazeer18a}
Noam Shazeer and Mitchell Stern.
\newblock Adafactor: Adaptive learning rates with sublinear memory cost.
\newblock In Jennifer Dy and Andreas Krause, editors, \emph{Proceedings of the
  35th International Conference on Machine Learning}, volume~80 of
  \emph{Proceedings of Machine Learning Research}, pages 4596--4604. PMLR,
  10--15 Jul 2018.
\newblock URL \url{https://proceedings.mlr.press/v80/shazeer18a.html}.

\bibitem[Shin et~al.(2021)Shin, Lin, Thomson, Chen, Roy, Platanios, Pauls,
  Klein, Eisner, and Van~Durme]{shin-etal-2021-constrained}
Richard Shin, Christopher Lin, Sam Thomson, Charles Chen, Subhro Roy,
  Emmanouil~Antonios Platanios, Adam Pauls, Dan Klein, Jason Eisner, and
  Benjamin Van~Durme.
\newblock Constrained language models yield few-shot semantic parsers.
\newblock In \emph{Proceedings of the 2021 Conference on Empirical Methods in
  Natural Language Processing}, pages 7699--7715, Online and Punta Cana,
  Dominican Republic, November 2021. Association for Computational Linguistics.
\newblock \doi{10.18653/v1/2021.emnlp-main.608}.
\newblock URL \url{https://aclanthology.org/2021.emnlp-main.608}.

\bibitem[Srivastava et~al.(2022)Srivastava, Rastogi, et~al.]{bigbench}
Aarohi Srivastava, Abhinav Rastogi, et~al.
\newblock Beyond the imitation game: Quantifying and extrapolating the
  capabilities of language models.
\newblock \emph{ArXiv}, abs/2206.04615, 2022.

\bibitem[Tian et~al.(2020)Tian, Song, Xia, and Zhang]{tian-etal-2020-improving}
Yuanhe Tian, Yan Song, Fei Xia, and Tong Zhang.
\newblock Improving constituency parsing with span attention.
\newblock In \emph{Findings of the Association for Computational Linguistics:
  EMNLP 2020}, pages 1691--1703, Online, November 2020. Association for
  Computational Linguistics.
\newblock \doi{10.18653/v1/2020.findings-emnlp.153}.
\newblock URL \url{https://aclanthology.org/2020.findings-emnlp.153}.

\bibitem[Touvron et~al.(2023)Touvron, Martin, Stone, Albert, Almahairi, Babaei,
  Bashlykov, Batra, Bhargava, Bhosale, Bikel, Blecher, Canton{-}Ferrer, Chen,
  Cucurull, Esiobu, Fernandes, Fu, Fu, Fuller, Gao, Goswami, Goyal, Hartshorn,
  Hosseini, Hou, Inan, Kardas, Kerkez, Khabsa, Kloumann, Korenev, Koura,
  Lachaux, Lavril, Lee, Liskovich, Lu, Mao, Martinet, Mihaylov, Mishra,
  Molybog, Nie, Poulton, Reizenstein, Rungta, Saladi, Schelten, Silva, Smith,
  Subramanian, Tan, Tang, Taylor, Williams, Kuan, Xu, Yan, Zarov, Zhang, Fan,
  Kambadur, Narang, Rodriguez, Stojnic, Edunov, and Scialom]{llama2}
Hugo Touvron, Louis Martin, Kevin Stone, Peter Albert, Amjad Almahairi, Yasmine
  Babaei, Nikolay Bashlykov, Soumya Batra, Prajjwal Bhargava, Shruti Bhosale,
  Dan Bikel, Lukas Blecher, Cristian Canton{-}Ferrer, Moya Chen, Guillem
  Cucurull, David Esiobu, Jude Fernandes, Jeremy Fu, Wenyin Fu, Brian Fuller,
  Cynthia Gao, Vedanuj Goswami, Naman Goyal, Anthony Hartshorn, Saghar
  Hosseini, Rui Hou, Hakan Inan, Marcin Kardas, Viktor Kerkez, Madian Khabsa,
  Isabel Kloumann, Artem Korenev, Punit~Singh Koura, Marie{-}Anne Lachaux,
  Thibaut Lavril, Jenya Lee, Diana Liskovich, Yinghai Lu, Yuning Mao, Xavier
  Martinet, Todor Mihaylov, Pushkar Mishra, Igor Molybog, Yixin Nie, Andrew
  Poulton, Jeremy Reizenstein, Rashi Rungta, Kalyan Saladi, Alan Schelten, Ruan
  Silva, Eric~Michael Smith, Ranjan Subramanian, Xiaoqing~Ellen Tan, Binh Tang,
  Ross Taylor, Adina Williams, Jian~Xiang Kuan, Puxin Xu, Zheng Yan, Iliyan
  Zarov, Yuchen Zhang, Angela Fan, Melanie Kambadur, Sharan Narang,
  Aur{\'{e}}lien Rodriguez, Robert Stojnic, Sergey Edunov, and Thomas Scialom.
\newblock Llama 2: Open foundation and fine-tuned chat models.
\newblock \emph{CoRR}, abs/2307.09288, 2023.
\newblock \doi{10.48550/arXiv.2307.09288}.
\newblock URL \url{https://doi.org/10.48550/arXiv.2307.09288}.

\bibitem[van Noord and Bos(2017{\natexlab{a}})]{van-noord-bos-2017-dealing}
Rik van Noord and Johan Bos.
\newblock Dealing with co-reference in neural semantic parsing.
\newblock In \emph{Proceedings of the 2nd Workshop on Semantic Deep Learning
  ({S}em{D}eep-2)}, pages 41--49, Montpellier, France, September
  2017{\natexlab{a}}. Association for Computational Linguistics.
\newblock URL \url{https://aclanthology.org/W17-7306}.

\bibitem[van Noord and Bos(2017{\natexlab{b}})]{van2017neural}
Rik van Noord and Johan Bos.
\newblock Neural semantic parsing by character-based translation: Experiments
  with abstract meaning representations.
\newblock \emph{Computational Linguistics in the Netherlands Journal},
  7:\penalty0 93--108, 2017{\natexlab{b}}.

\bibitem[Wang et~al.(2018)Wang, Singh, Michael, Hill, Levy, and
  Bowman]{wang-etal-2018-glue}
Alex Wang, Amanpreet Singh, Julian Michael, Felix Hill, Omer Levy, and Samuel
  Bowman.
\newblock {GLUE}: A multi-task benchmark and analysis platform for natural
  language understanding.
\newblock In \emph{Proceedings of the 2018 {EMNLP} Workshop {B}lackbox{NLP}:
  Analyzing and Interpreting Neural Networks for {NLP}}, pages 353--355,
  Brussels, Belgium, November 2018. Association for Computational Linguistics.
\newblock \doi{10.18653/v1/W18-5446}.
\newblock URL \url{https://aclanthology.org/W18-5446}.

\bibitem[Wang et~al.(2019)Wang, Pruksachatkun, Nangia, Singh, Michael, Hill,
  Levy, and Bowman]{wang-etal-2019-superglue}
Alex Wang, Yada Pruksachatkun, Nikita Nangia, Amanpreet Singh, Julian Michael,
  Felix Hill, Omer Levy, and Samuel Bowman.
\newblock Superglue: A stickier benchmark for general-purpose language
  understanding systems.
\newblock In H.~Wallach, H.~Larochelle, A.~Beygelzimer, F.~d\textquotesingle
  Alch\'{e}-Buc, E.~Fox, and R.~Garnett, editors, \emph{Advances in Neural
  Information Processing Systems}, volume~32. Curran Associates, Inc., 2019.
\newblock URL
  \url{https://proceedings.neurips.cc/paper/2019/file/4496bf24afe7fab6f046bf4923da8de6-Paper.pdf}.

\bibitem[Wang et~al.(2021{\natexlab{a}})Wang, Wang, Joty, and
  Hoi]{wang-etal-2021-codet5}
Yue Wang, Weishi Wang, Shafiq Joty, and Steven~C.H. Hoi.
\newblock {C}ode{T}5: Identifier-aware unified pre-trained encoder-decoder
  models for code understanding and generation.
\newblock In \emph{Proceedings of the 2021 Conference on Empirical Methods in
  Natural Language Processing}, pages 8696--8708, Online and Punta Cana,
  Dominican Republic, November 2021{\natexlab{a}}. Association for
  Computational Linguistics.
\newblock \doi{10.18653/v1/2021.emnlp-main.685}.
\newblock URL \url{https://aclanthology.org/2021.emnlp-main.685}.

\bibitem[Wang et~al.(2015)Wang, Berant, and Liang]{wang-etal-2015-building}
Yushi Wang, Jonathan Berant, and Percy Liang.
\newblock Building a semantic parser overnight.
\newblock In \emph{Proceedings of the 53rd Annual Meeting of the Association
  for Computational Linguistics and the 7th International Joint Conference on
  Natural Language Processing (Volume 1: Long Papers)}, pages 1332--1342,
  Beijing, China, July 2015.
\newblock \doi{10.3115/v1/P15-1129}.
\newblock URL \url{https://www.aclweb.org/anthology/P15-1129}.

\bibitem[Wang et~al.(2021{\natexlab{b}})Wang, Yu, Firat, and
  Cao]{zero-label-wang-2021}
Zirui Wang, Adams~Wei Yu, Orhan Firat, and Yuan Cao.
\newblock Towards zero-label language learning.
\newblock \emph{CoRR}, abs/2109.09193, 2021{\natexlab{b}}.
\newblock URL \url{https://arxiv.org/abs/2109.09193}.

\bibitem[Wiseman and Rush(2016)]{wiseman-rush-2016-sequence}
Sam Wiseman and Alexander~M. Rush.
\newblock Sequence-to-sequence learning as beam-search optimization.
\newblock In \emph{Proceedings of the 2016 Conference on Empirical Methods in
  Natural Language Processing}, pages 1296--1306, Austin, Texas, November 2016.
  Association for Computational Linguistics.
\newblock \doi{10.18653/v1/D16-1137}.
\newblock URL \url{https://aclanthology.org/D16-1137}.

\bibitem[Xie et~al.(2022)Xie, Wu, Shi, Zhong, Scholak, Yasunaga, Wu, Zhong,
  Yin, Wang, Zhong, Wang, Li, Boyle, Ni, Yao, Radev, Xiong, Kong, Zhang, Smith,
  Zettlemoyer, and Yu]{UnifiedSKG}
Tianbao Xie, Chen~Henry Wu, Peng Shi, Ruiqi Zhong, Torsten Scholak, Michihiro
  Yasunaga, Chien-Sheng Wu, Ming Zhong, Pengcheng Yin, Sida~I. Wang, Victor
  Zhong, Bailin Wang, Chengzu Li, Connor Boyle, Ansong Ni, Ziyu Yao, Dragomir
  Radev, Caiming Xiong, Lingpeng Kong, Rui Zhang, Noah~A. Smith, Luke
  Zettlemoyer, and Tao Yu.
\newblock Unifiedskg: Unifying and multi-tasking structured knowledge grounding
  with text-to-text language models.
\newblock \emph{EMNLP}, 2022.

\bibitem[Yang and Tu(2022)]{yang-tu-2022-bottom}
Songlin Yang and Kewei Tu.
\newblock Bottom-up constituency parsing and nested named entity recognition
  with pointer networks.
\newblock In \emph{Proceedings of the 60th Annual Meeting of the Association
  for Computational Linguistics (Volume 1: Long Papers)}, pages 2403--2416,
  Dublin, Ireland, May 2022. Association for Computational Linguistics.
\newblock \doi{10.18653/v1/2022.acl-long.171}.
\newblock URL \url{https://aclanthology.org/2022.acl-long.171}.

\bibitem[Yu et~al.(2018)Yu, Zhang, Yang, Yasunaga, Wang, Li, Ma, Li, Yao,
  Roman, Zhang, and Radev]{yu-etal-2018-spider}
Tao Yu, Rui Zhang, Kai Yang, Michihiro Yasunaga, Dongxu Wang, Zifan Li, James
  Ma, Irene Li, Qingning Yao, Shanelle Roman, Zilin Zhang, and Dragomir Radev.
\newblock {S}pider: A large-scale human-labeled dataset for complex and
  cross-domain semantic parsing and text-to-{SQL} task.
\newblock In \emph{Proceedings of the 2018 Conference on Empirical Methods in
  Natural Language Processing}, pages 3911--3921, Brussels, Belgium,
  October-November 2018.
\newblock \doi{10.18653/v1/D18-1425}.
\newblock URL \url{https://www.aclweb.org/anthology/D18-1425}.

\bibitem[Yu et~al.(2019)Yu, Zhang, Er, Li, Xue, Pang, Lin, Tan, Shi, Li, Jiang,
  Yasunaga, Shim, Chen, Fabbri, Li, Chen, Zhang, Dixit, Zhang, Xiong, Socher,
  Lasecki, and Radev]{yu-etal-2019-cosql}
Tao Yu, Rui Zhang, Heyang Er, Suyi Li, Eric Xue, Bo~Pang, Xi~Victoria Lin,
  Yi~Chern Tan, Tianze Shi, Zihan Li, Youxuan Jiang, Michihiro Yasunaga,
  Sungrok Shim, Tao Chen, Alexander Fabbri, Zifan Li, Luyao Chen, Yuwen Zhang,
  Shreya Dixit, Vincent Zhang, Caiming Xiong, Richard Socher, Walter Lasecki,
  and Dragomir Radev.
\newblock {C}o{SQL}: A conversational text-to-{SQL} challenge towards
  cross-domain natural language interfaces to databases.
\newblock In \emph{Proceedings of the 2019 Conference on Empirical Methods in
  Natural Language Processing and the 9th International Joint Conference on
  Natural Language Processing (EMNLP-IJCNLP)}, pages 1962--1979, Hong Kong,
  China, November 2019. Association for Computational Linguistics.
\newblock \doi{10.18653/v1/D19-1204}.
\newblock URL \url{https://aclanthology.org/D19-1204}.

\bibitem[Zeman et~al.(2022)]{univ-deps-211}
Daniel Zeman et~al.
\newblock Universal dependencies 2.11, 2022.
\newblock URL \url{http://hdl.handle.net/11234/1-4923}.
\newblock {LINDAT}/{CLARIAH}-{CZ} digital library at the Institute of Formal
  and Applied Linguistics ({{\'U}FAL}), Faculty of Mathematics and Physics,
  Charles University.

\bibitem[Zhang et~al.(2017)Zhang, Liu, Li, Zhou, and
  Chen]{zhang-etal-2017-stack}
Zhirui Zhang, Shujie Liu, Mu~Li, Ming Zhou, and Enhong Chen.
\newblock Stack-based multi-layer attention for transition-based dependency
  parsing.
\newblock In \emph{Proceedings of the 2017 Conference on Empirical Methods in
  Natural Language Processing}, pages 1677--1682, Copenhagen, Denmark,
  September 2017. Association for Computational Linguistics.
\newblock \doi{10.18653/v1/D17-1175}.
\newblock URL \url{https://aclanthology.org/D17-1175}.

\bibitem[Zhong et~al.(2020)Zhong, Yu, and Klein]{ruiqi20}
Ruiqi Zhong, Tao Yu, and Dan Klein.
\newblock Semantic evaluation for text-to-sql with distilled test suite.
\newblock In \emph{The 2020 Conference on Empirical Methods in Natural Language
  Processing}. Association for Computational Linguistics, 2020.

\end{thebibliography}


\newpage
\appendix
\section{Variance of Results}
\label{sec:variance}
All low-resource results are a mean of three training data splits. Table
\ref{tab:variance} reports the average standard deviation for each model over the three low resource splits. We find a high standard deviation of GPT-3 and Codex; one of
the factors being the small size of the test set 
(\num{100} examples). Fine-tuned models show relatively low variance, consistently having standard deviation lower than $2\%$.
\begin{table}[!htbp]
\caption{Standard deviation of the scores for each language model over the three
low resource splits.}
\label{tab:variance}
\small
\centering
\begin{tabular}{@{}lr@{}}
\toprule
{\bf Model} & {\bf Avg. Standard Deviation} \\\midrule
 GPT-3 & 4.7 \\
 Codex & 3.2 \\
\addlinespace
 T5-base & 1.2 \\
 CodeT5-base & 1.1 \\
 BART-large & 1.4 \\
 T5-large & 2.0 \\
 T5-3B & 1.5 \\
\bottomrule
\end{tabular}
\end{table}

\section{Compute Details}
\label{sec:compute}
We used Microsoft Azure to run all our experiments. For experiments with T5-base,
CodeT5-base and BART-large, we used a single V100 GPU with $32$ GB memory. We used
$2$ V100 GPUs for fine-tuning T5-large and $4$ GPUs for fine-tuning T5-3B. Training time
ranged from $2$ to $6$ hours depending on the size of the model.

\section{Dataset and Model License and Privacy Implications}
\label{sec:licence}
The licenses of the datasets included in BenchCLAMP are listed in Table \ref{tab:licence}.
We could not find a license for MTOP, but the dataset was made freely available by
the creators. Our benchmark is available under the MIT licence at {\small\url{https://github.com/microsoft/semantic_parsing_with_constrained_lm}}. 

The datasets in BenchCLAMP either use sentences from news articles or utterances
generated by crowd-workers following instructions to simulate user interactions. No real
user interaction data was used to generate these datasets. As a result, there is no
privacy risk in using these datasets.

For fine-tuning experiments, we use pre-trained models from Huggingface. All models 
were released under the Apache 2.0 license.
\begin{table}[H]
\label{tab:licence}
\caption{Licenses of datasets covered by BenchCLAMP.}
\small
\centering
\begin{tabular}{@{}l r}
\toprule
\bf Dataset & \bf License \\
\midrule
SMCalFlow \cite{andreas-etal-2020-task} & CC BY-SA 4.0 \\
TreeDST \cite{cheng-etal-2020-conversational} & CC BY-SA 3.0 \\
MTOP \cite{mtop-li-2021} & --\\
Overnight \cite{wang-etal-2015-building} & CC BY-SA 4.0 \\
Spider \cite{yu-etal-2018-spider} & CC BY-SA 4.0 \\
CoSQL \cite{yu-etal-2019-cosql} & CC BY-SA 4.0 \\
AMR 2.0 \cite{banarescu-etal-2013-AMR} & LDC License\\
PTB 3 \cite{marcus-etal-1993-ptb} (Constituency parsing) & LDC License\\
UD-EWT \cite{univ-deps-211} (Dependency parsing) & CC BY-SA 4.0 International\\
\bottomrule
\end{tabular}
\end{table}

\end{document}